\documentclass[10pt,twocolumn,letterpaper]{article}

\usepackage{iccv}
\usepackage{times}
\usepackage{epsfig}
\usepackage{graphicx}
\usepackage{amsmath}
\usepackage{amssymb}

\usepackage{bbm}
\usepackage{multirow}
\usepackage{makecell}
\usepackage{enumitem}

\usepackage{hyperref}
\hypersetup{colorlinks,allcolors=black}

\usepackage{color}

\def\Approach{ASBU}

\def\block1{{$\sf block1$}}
\def\block2{{$\sf block2$}}
\def\block3{{$\sf block3$}}
\def\block4{{$\sf block4$}}

\def\m{{\bf m}}

\iccvfinalcopy 

\def\iccvPaperID{4365} 

\ificcvfinal\fi

\begin{document}

\title{A Weakly Supervised Amodal Segmenter with Boundary Uncertainty Estimation }
\author{Khoi Nguyen and Sinisa Todorovic\\
Oregon State University\\
Corvallis, OR 97330, USA\\
{\tt\small {\{nguyenkh,sinisa\}}@oregonstate.edu}
}

\maketitle
\ificcvfinal\thispagestyle{empty}\fi

\begin{abstract}
This paper addresses weakly supervised amodal instance segmentation, where the goal is to segment both visible and occluded (amodal) object parts, while training provides only ground-truth visible (modal) segmentations. Following prior work, we use data manipulation to generate occlusions in training images and thus train a segmenter to predict amodal segmentations of the manipulated data. The resulting predictions on training images are taken as the pseudo-ground truth for the standard training of Mask-RCNN, which we use for amodal instance segmentation of test images. For generating the pseudo-ground truth, we specify a new  Amodal Segmenter based on Boundary Uncertainty estimation (\Approach) and make two contributions. First, while prior work uses the occluder's mask, our \Approach~uses the occlusion boundary as input.
Second,  \Approach~ estimates an uncertainty map of the prediction. The estimated uncertainty regularizes learning such that lower segmentation loss is incurred on regions with high uncertainty.  \Approach~achieves significant performance improvement relative to the state of the art on the COCOA and KINS datasets in three tasks: amodal instance segmentation, amodal completion, and ordering recovery.
\end{abstract}

\section{Introduction}\label{sec:intro}

In this paper, we seek to address the problem of weakly supervised amodal instance segmentation (WAIS). Our goal is to segment both visible and occluded (amodal) parts of object instances in images. The weak supervision in training provides only ground-truth visible (modal) instance segmentations. Important applications of amodal segmentation include autonomous driving and robot path planning, where identifying the whole spatial extents of partially occluded objects is critical. Considering this problem under weak supervision is also important 
because human annotators often cannot provide reliable ground truth. For example, different annotators are likely to have very different and sometimes poor guesses of occluded object parts.

There is scant prior work on WAIS.
Following recent PCNet  \cite{zhan2020self}, our training consists of two stages. First, we use data augmentation to train a common image segmenter -- UNet \cite{ronneberger2015u} -- on manipulated training images to predict their amodal segmentations. As input to UNet, we use the available ground-truth modal segmentation and information about where the data augmentation generated the occlusion in the training image. In the second training stage, UNet's amodal segmentations
are taken as a pseudo-ground truth for learning a standard instance segmenter -- Mask-RCNN \cite{he2017mask}, as in \cite{zhan2020self}.  On test images with occlusions, Mask-RCNN trained on the pseudo-ground truth is expected to output correct amodal instance segmentation. 

Our contributions are aimed at advancing the first training stage, and include: (1) a new way to exploit the weak supervision for training of  UNet; and (2) enabling UNet to estimate uncertainty of the predicted amodal segmentation, and enforcing the training of UNet to explicitly minimize this uncertainty.

\begin{figure*}[h!]
    \centering
    \includegraphics[scale=0.34]{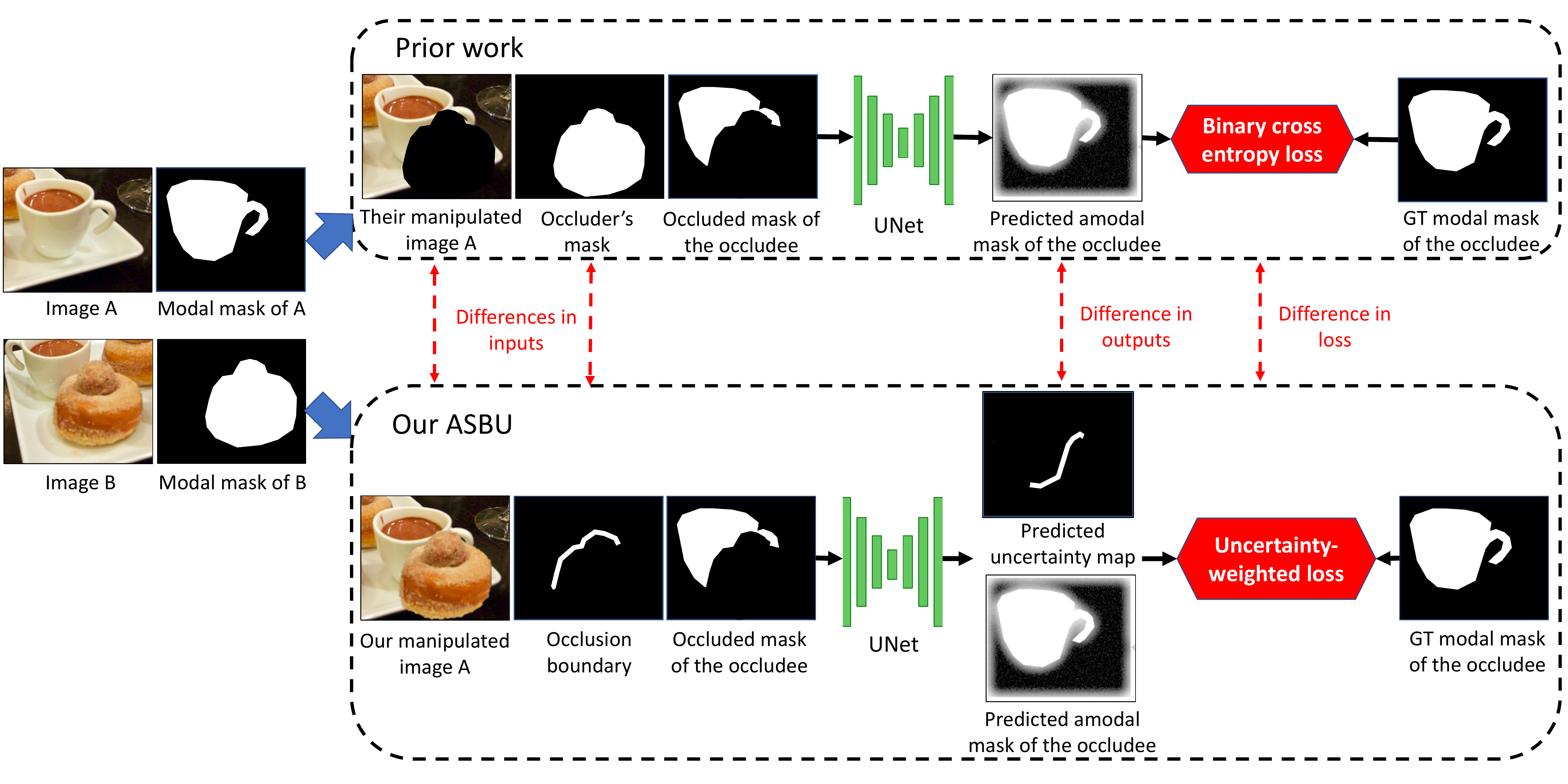}
    \caption{(Top) A recent approach to WAIS \cite{zhan2020self} where UNet \cite{ronneberger2015u} is trained with binary cross-entropy loss to predict amodal segmentation from the manipulated training image, the occluder's mask, and the occluded mask of the occludee. (Bottom) Our approach, called \Approach, differs \cite{zhan2020self} in terms of input, output, and loss. For input, instead of the occluder's mask, we use the occlusion boundary, and instead of ``zeros'' we use matting to superimpose the occluder onto the image. This means that we remove the information about the occluder's spatial extent from the input to UNet. For output, along with amodal segmentation, we additionally estimate uncertainty of prediction. For loss, we use uncertainty to appropriately weight loss. GT stands for ground truth.}
    \label{fig:data_manipulation}
\end{figure*}

Our first contribution is motivated by the following limitation of PCNet \cite{zhan2020self}. For manipulating training images, as illustrated in Fig.~\ref{fig:data_manipulation}, PCNet randomly places an {\em occluder} object onto an {\em occludee} object based on their ground-truth modal segmentation masks, and in this way artificially generates the occluded mask of the occludee. Then, as three inputs to UNet, PCNet uses the manipulated training image, the occluded mask of the occludee, and the occluder's mask. However, using the occluder's mask as input to UNet puts the restrictive constraint that the occluder itself cannot be occluded by another object. To address this limitation,  PCNet estimates an object ordering graph in the image, and for the occluder selects a union of all objects estimated as closer to the camera than the occludee (i.e., a union of multiple occluders). Our novelty is in replacing the occluder's mask with the occlusion boundary in the input to UNet, as depicted in Fig.~\ref{fig:data_manipulation}. Thus, our occluders are allowed to be themselves partially occluded by some other objects. This reduces complexity of estimating the pseudo-ground-truth amodal segmentation, as we do not need to estimate the object ordering graph.

\begin{figure*}[h!]
    \centering
    \includegraphics[scale=0.30]{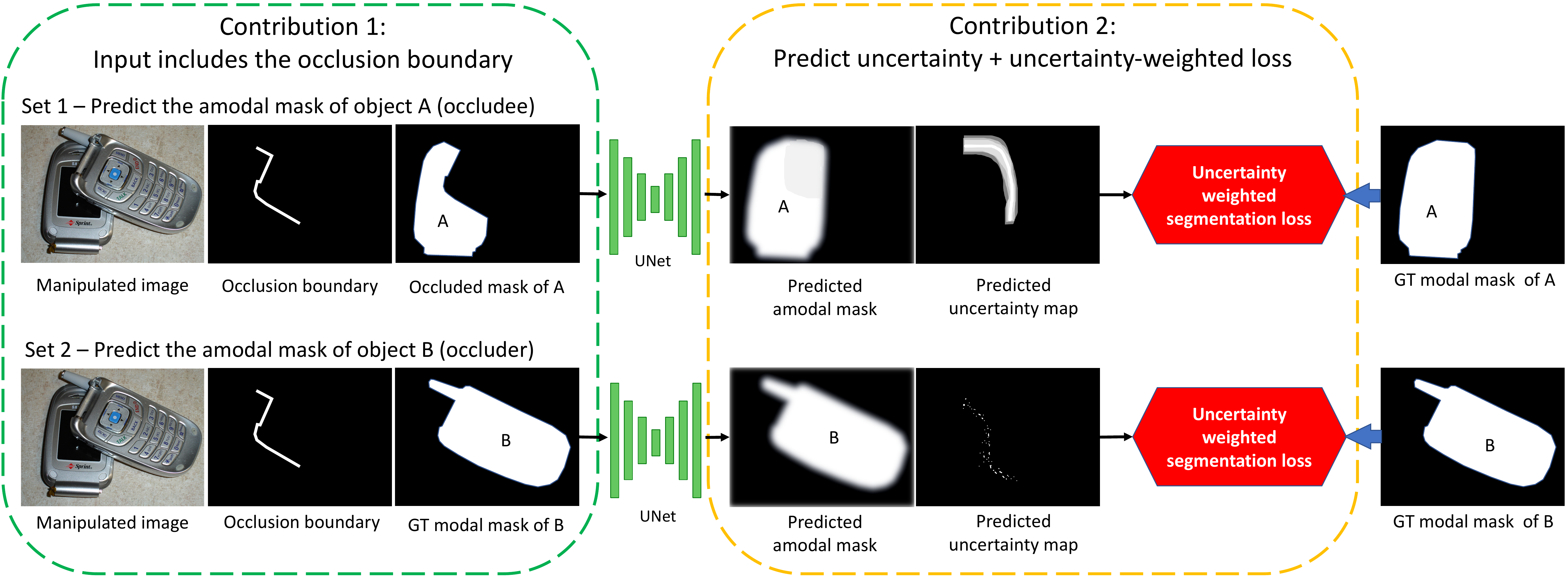}
    \caption{ \Approach~is trained on two sets of input triplets so as to learn when to spatially extend and when not to extend the input segmentation mask in prediction. Our first contribution is in the input to \Approach -- namely, we replace
     the occluder mask used in \cite{zhan2020self} with the occlusion boundary mask. Our second contribution is in the prediction of the uncertainty map and using uncertainty to appropriately weight loss. The figure shows that the predicted uncertainty is usually low on regions close to the occlusion. This is used for regularizing learning. Note that we adjusted brightness for visualizing uncertainty (the brighter pixels in the uncertainty map the higher uncertainty), because it is significantly lower for set 2 than for set 1.}
    \label{fig:main_diagram}
\end{figure*}

Our second contribution is aimed at accounting for shape priors. As shown in Fig.~\ref{fig:main_diagram}, we {\em implicitly} capture a ``shape prior'' through learning to estimate an uncertainty map for the predicted amodal segmentation. In our experiments, we observe that the estimated uncertainty typically takes low (high) values over regions far away (close) to the occlusion boundary.  This suggests that our uncertainty map is capable of representing a spatial distribution of object shapes, and hence can be used for regularizing our learning.  Our regularization uses the estimated uncertainty map to appropriately modulate a difference between the predicted amodal segmentation and the original ground-truth mask of the occludee (before the occlusion), such that lower loss is incurred on regions with high uncertainty.

Our two contributions are incorporated in the new  Amodal Segmenter with Boundary Uncertainty estimation (\Approach).  \Approach~is evaluated on the COCOA \cite{zhu2017semantic} and KINS  \cite{qi2019amodal} datasets on three tasks: amodal instance segmentation, amodal completion, and ordering recovery. \Approach~significantly outperforms the state of the art in all three tasks.

In the following, Sec.~\ref{sec:related_work} reviews previous work; Sec.~\ref{sec:method} specifies \Approach; Sec.~\ref{sec:loss} formalizes our uncertainty estimation and uncertainty weighted loss;
Sec.~\ref{sec:experiments} presents our implementation details and experimental results; and Sec.~\ref{sec:conclusions} concludes the paper.

\section{Related Work}
\label{sec:related_work}
This section reviews closely related work.

\textbf{Instance Segmentation} is aimed at labeling pixels with object instance labels, and can be addressed with \textit{bounding-box-based} and \textit{bounding-box-free} methods. In the former \cite{he2017mask,li2017fully,chen2018masklab,liu2018path,bolya2019yolact}, for every detected bounding box, a foreground object is segmented. In the latter \cite{kirillov2017instancecut,kong2018recurrent,chandra2017dense,fathi2017semantic}, first, a semantic segmentation is obtained, and then pixels of the same semantic class are clustered into instances based on visual cues such as object center or inner sign distance function. 
All of these approaches segment only visible object parts and thus are not suitable for our problem.

\textbf{Amodal Instance Segmentation} infers visible and occluded object parts, under full supervision in training. For the ground truth, prior work uses amodal segmentation of either real images \cite{li2016amodal, zhu2017semantic, qi2019amodal, follmann2019learning, ke2021deep} or synthetic data \cite{ehsani2018segan,hu2019sail,kihara2016shadows}. However, the existing quality of synthetic data introduces a domain gap between training on synthetic images and testing on real images, resulting in a considerable performance difference between the two domains.

\textbf{Amodal Instance Completion} differs from amodal instance segmentation since the goal is to complete occluded parts of an object given its modal mask, whereas in amodal instance segmentation the modal mask is not provided. Prior work typically uses the Gestalt principles and makes certain assumptions about shape convexity and length. For example, amodal instance completion has been addressed by using Euler spiral, cubic Bezier curves, shape primitives, and variational auto-encoder in \cite{kimia2003euler,lin2016computational,silberman2014contour, amodalVAE20}. Our first stage of training for generating the amodal pseudo-ground truth is based on amodal instance completion. We evaluate \Approach~on the task of amodal instance completion.

\begin{figure}[h!]
    \centering
    \includegraphics[scale=0.45]{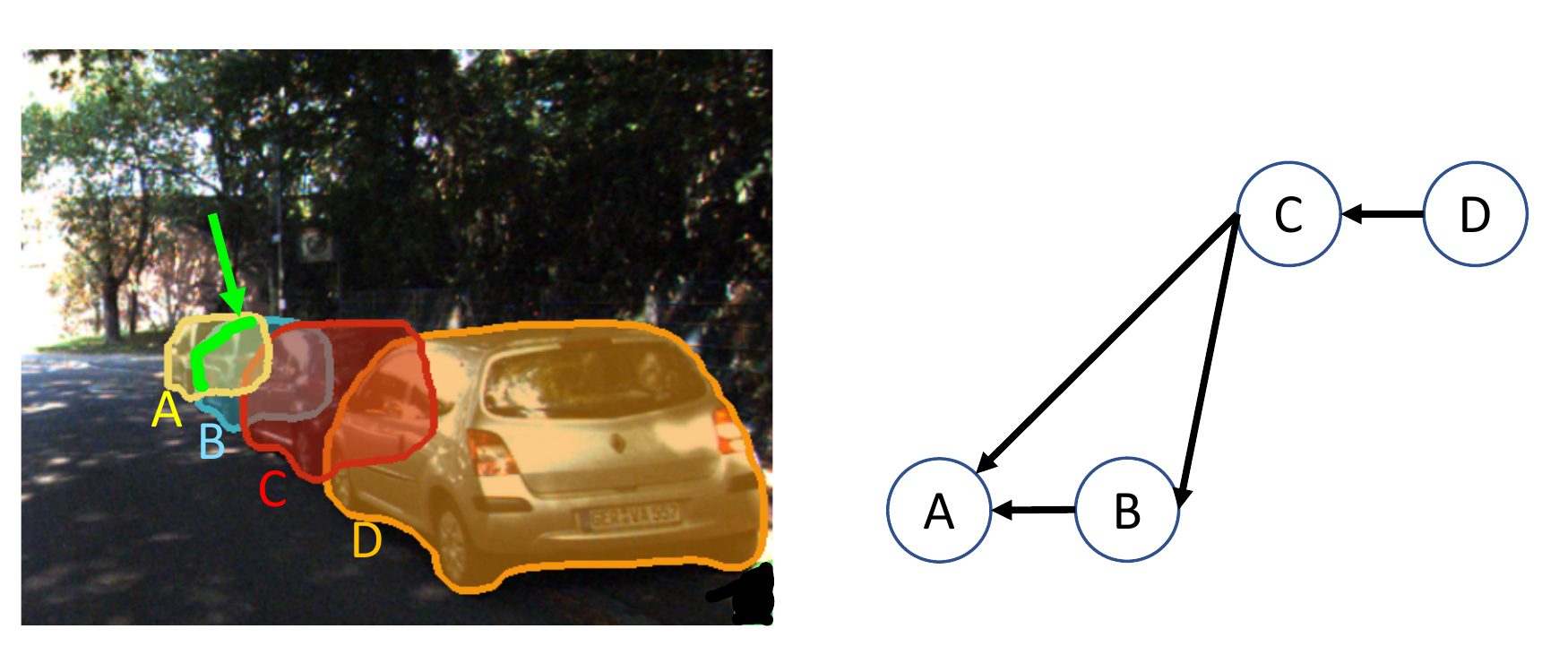}
    \caption{ 
An example of a occluder partially occluded by other objects. A is occluded by B and C, B is occluded by C, and C is occluded by D. For input in the first stage of our training, we use only the occlusion boundary (green), whereas \cite{zhan2020self} first estimates the ordering graph of A, B, C, and D (on the right), and then takes a union mask of B, C, and D as input.}
    \label{fig:cars}
\end{figure}

\begin{figure}[h!]
    \centering
    \includegraphics[scale=0.45]{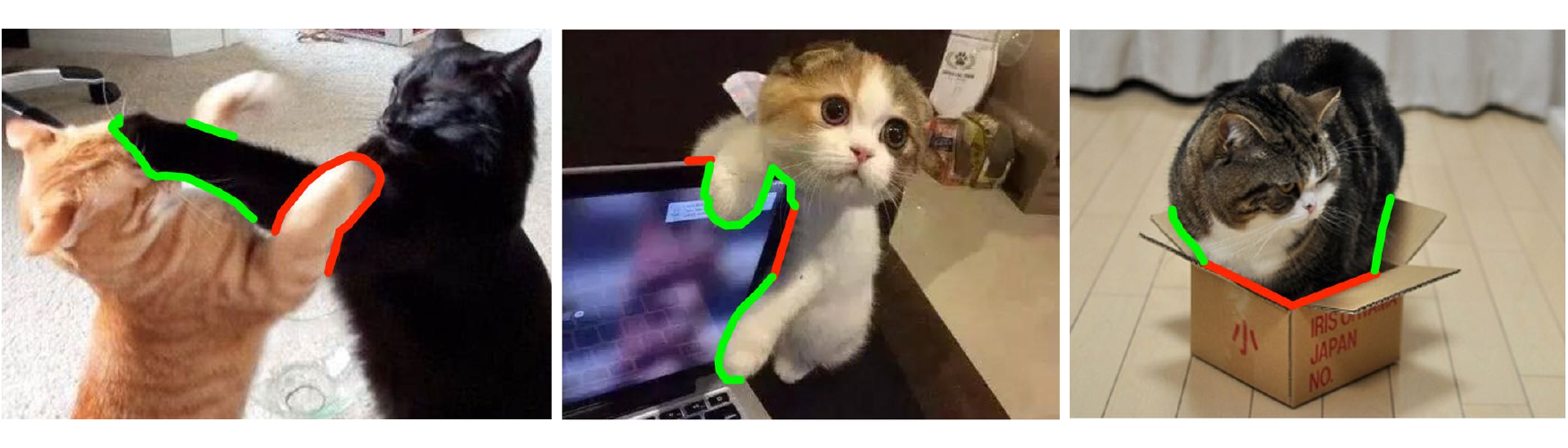}
    \caption{Examples of entangled objects occluding each other, where the occlusion relationships cannot be uniquely represented by an ordering graph, and thus are very challenging for PCNet \cite{zhan2020self}. As shown in Fig.~\ref{fig:mutual_occlusion}, we successfully address these cases.}
    \label{fig:cats}
\end{figure}

\textbf{WAIS} provides access only to modal-mask annotations in training. 
Recent work \cite{zhan2020self} begins by converting modal-mask annotations of training images into pseudo amodal masks in a self-supervised manner, as illustrated in Fig.~\ref{fig:data_manipulation}. However, in a complex scene, the occluder can also be occluded by another object, as shown in Fig.~\ref{fig:cars}, which requires \cite{zhan2020self} to construct an object ordering graph. This increases complexity of their first stage of training of UNet, and is not even suitable for addressing cases of entangled partial occlusions when the occluder-occludee relationship of a pair of objects is not unique, as illustrated in Fig.~\ref{fig:cats}. We overcome this limitation by replacing the occluder's mask with the occlusion boundary for our input. Unlike \cite{zhan2020self}, we effectively  remove from our input any information about the occluder's spatial extent. Consequently, we do not need to estimate the object ordering graph as in \cite{zhan2020self}.

Recently, \cite{sun2020weakly} proposes to learn shape priors for each category by using modal object bounding boxes. Then the amodal object bounding box is obtained by aligning modal box with learned shape priors. This approach only works well with low-deformation object categories such as car and motorbike of KINS \cite{qi2019amodal} so that we can robustly learn object shape priors. In contrast, \Approach~can handle many types of object category as in COCOA \cite{zhu2017semantic}.


\textbf{Uncertainty Estimation in Segmentation} has a long track record in the literature. Prior work
typically estimates aleatoric uncertainty (data uncertainty) \cite{kohl2018probabilistic} and epistemic uncertainty (weight uncertainty) \cite{kendall2015bayesian}, where the former estimates noise in observations and the latter accounts for a distribution of model parameters. For example, in \cite{kohl2018probabilistic}, UNet \cite{ronneberger2015unet} is extended with a variational auto-encoder for aleatoric uncertainty estimation. In \cite{kendall2015bayesian}, estimation of a distribution of the SegNet parameters \cite{badrinarayanan2017segnet} replaces the common fixed-point parameter estimation. Shape priors have also been studied in the following related work \cite{kihara2016shadows, kimia2003euler, lin2016computational, silberman2014contour}.


\section{Our Approach}
\label{sec:method}

The section specifies our \Approach. Fig.~\ref{fig:main_diagram} shows that \Approach~uses two distinct sets of inputs for training of UNet to jointly predict the amodal segmentation mask and the associated uncertainty map. These predictions incur an uncertainty-weighted loss function, specified such that our training minimizes both uncertainty and errors in the predicted amodal segmentation.

Our data manipulation of training images, first, randomly samples two objects as occludee and occluder, then, randomly samples a relative position between their modal masks such that the occluder's mask partially occludes the occludee's mask, and finally prepares the following two sets of input data:
\begin{enumerate}[itemsep=-1pt,topsep=1pt, leftmargin=20pt]
    \item (set 1 and set 2) Manipulated training image, where the occluder's image is superimposed onto the occludee's image with matting for  realistic appearance;
    \item (set 1 and set 2) Occlusion boundary mask, estimated as an intersection of the morphologically enlarged masks of the occludee and occluder;
    \item  \begin{enumerate}[itemsep=1pt,topsep=-2pt, leftmargin=30pt]
    \item[(set 1)]  Occluded mask of the occludee, where pixels of the ground-truth modal mask of the occludee covered by the occluder are zero.
    \item[(set 2)]  Ground-truth modal mask of the occluder.
\end{enumerate}
 \end{enumerate} %

 \vspace{5pt}
 
Importantly, while both input sets are used in training, \Approach~is not aware if the the input segmentation mask comes from the occludee or from the occluder.  In this way, \Approach~is trained to identify when and how to perform amodal completion of the input segmentation mask. Specifically, for set~1 at the input, \Approach~is supposed to learn {\em to extend} the input segmentation mask in the region with zero pixels, which is delineated by the input occlusion boundary, because this region is likely to represent the manipulated occlusion. On the other hand, for set~2 at the input, \Approach~is supposed to learn {\em not to extend} the input segmentation mask in the zero-valued region.

\section{Uncertainty Weighted Segmentation Loss}
\label{sec:loss}

Our uncertainty estimation is aimed at implicitly capturing a ``shape prior'' of training object instances, which is used for regularizing our  learning. Thus, our learning has the following two objectives: 
\begin{enumerate}[itemsep=-1pt,topsep=1pt, leftmargin=15pt]
\item Minimizing uncertainty of the predicted amodal segmentation, so when uncertainty is estimated as large it captures a truly large variability in plausible shapes; 
\item Penalizing a difference between the predicted amodal segmentation and the ground truth in an adaptive manner, such that this loss is appropriately reduced when the shape is estimated to come from a highly variable distribution -- i.e., when the estimated uncertainty is high. 
\end{enumerate}

To this end, our \Approach~extends UNet to output a $H{\times}W{\times}2$ feature map which has two channels, one for amodal segmentation prediction and the another for the uncertainty prediction, where $H$ and $W$ denote the height and width of the input image. The predicted values for amodal segmentation, $\{m_i:i=1,\dots,N\}$, $N=H\cdot W$, are output by the sigmoid function, so they range in $m_i\in[0,1]$. The predicted  uncertainty values, $\{u_i:i=1,\dots,N\}$, are output by the softplus function, $\text{softplus}(z) = \log (1+\exp(z))$, so they are positive  $u_i\in\mathbbm{R}^+$.

For learning to jointly predict $\{m_i\}$ and $\{u_i\}$, we specify a new uncertainty-weighted segmentation loss. The prediction of $\{m_i\}$ can be supervised by the corresponding original modal mask as it was before the data manipulation $\{m_i^*\}$. On the other hand, the prediction of $\{u_i\}$ remains unsupervised, since there is no ground-truth annotation of uncertainty in our training data. We integrate the mentioned supervised and unsupervised training strategies in the following loss:%
\begin{align}
    L  =& \displaystyle\frac{1}{N} \sum_{i=1}^{N} \mathbbm{1}(m_i^{C}=0) L_i + \lambda \mathbbm{1}(m_i^{C}=1) L_i, \nonumber \\
    L_i=& \displaystyle\frac{1}{2} \left[ \left(\frac{m^*_i - m_i}{u_i}\right)^2 + u_i^2 \right],
\label{eq:loss_function}
\end{align} 
where $m^C$ is the mask of the occluder, $\lambda$ is a positive constant, $\mathbbm{1}(\cdot)$ is the indicator function, and $N$ is the number of pixels in the image. We empirically estimate that $\lambda=5$ gives the best results, i.e., we put more weight on predictions inside the occluder mask.

Our loss in \eqref{eq:loss_function} is inspired by the Mumford-Shah energy \cite{mumford1989optimal}, whose minimization is a classical framework for image segmentation. The Mumford-Shah energy has two terms -- namely, data term and regularization term. By minimizing a sum of these two terms, the predicted segmentation is encouraged to simultaneously minimize energy and complexity (e.g., favor solutions with a high log-likelihood and smooth boundaries). 

Similarly, our loss $L_i$ in \eqref{eq:loss_function} also has two terms. The first term can be interpreted as the data term for minimizing the energy of a weighted difference between the predicted amodal segmentation $m_i$ and ground truth $m_i^*$. The weighting is inversely proportional to the estimated uncertainty for pixels $i$, such that a lower loss is backpropagated in our training for pixels with higher uncertainty. Since $u_i$ is typically higher on object boundaries than over other object parts, the specified weighting effectively accounts for shape variability. The second term in $L_i$ can be interpreted as the regularization term for penalizing large $u_i$ values. In our experiments, we observe that this regularization favors zero $u_i$ values over object regions that are not close to the boundary. Consequently, the data term in $L_i$ for such regions introduces a large loss for any errors in the amodal segmentation $m_i$, because $u_i\approx 0$.  

Our loss formulation fundamentally differs from other recent approaches aimed at estimating uncertainty for object segmentation.  For example, in \cite{kendall2017uncertainties}, segmentation and its uncertainty are assumed as governed by a Gaussian distribution with the following loss function:
\begin{equation}
    L_i^{\text{Gaussian}}= \frac{1}{2} \left [ \frac{(m^*_i - m_i)^2}{u^2_i} + \log u^2_i \right ] 
\label{eq:uncertainty_gaussian}
\end{equation}
In contrast, we do not explicitly specify any probability distribution of box locations. Also, unlike our $L_i$ in \eqref{eq:loss_function},  $L_i^{\text{Gaussian}}$ in \eqref{eq:uncertainty_gaussian} minimizes uncertainty only when $u > 1$. In \cite{amodalVAE20}, uncertainty is modeled in the latent space of their variational auto-encoder, and the amodal mask is predicted by sampling multiple latent codes from a  Gaussian distribution. In contrast, we predict and regularize our uncertainty map directly in the spatial domain.



\section{Results}
\label{sec:experiments}

\begin{table*}[]
\begin{center}
\small
\begin{tabular}{c|cc|cc|ccc}
\hline
\hline
\multirow{2}{*}{\textbf{Methods}} & \multicolumn{2}{c|}{\textbf{COCOA-val}} & \multicolumn{2}{c|}{\textbf{COCOA-test}} & \multicolumn{3}{c}{\textbf{KINS-test}} \\ \cline{2-8}
 & \textbf{O-Acc} & \textbf{mIOU} & \textbf{O-Acc} & \textbf{mIoU} & \textbf{O-Acc} & \textbf{mIoU} & \textbf{inv-mIoU}  \\ \hline
Amodal-VAE \cite{amodalVAE20} (reported)      & - & - & - & - & - & 94.68 & 62.85 \\
PCNet-m \cite{zhan2020self} (reported)        & 87.10 & 81.35 & -     & -     & 92.50 & 94.76 & -  \\
PCNet-m (reproduced)                          & 85.75 & 80.73 & 86.73 & 86.63 & 91.73 & 94.52 & 59.24 \\
\hline
\mbox{Boundary$\to$PCNet-m}                   & 89.01 & 82.85 & 89.22 & 88.67 & 92.26 & 94.65 & 62.77 \\
\mbox{Uncertainty$\to$PCNet-m}                & 88.60 & 82.49 & 88.40 & 88.15 & 92.08 & 94.61 & 62.00 \\
\mbox{uBCE$\to$\Approach}                     & 89.23 & 83.18 & 89.32 & 88.10 & 92.15 & 94.34 & 63.41 \\
\hline
\Approach                                     & \textbf{90.33} & \textbf{84.22} & \textbf{90.77} & \textbf{89.87} & \textbf{92.65} & \textbf{94.83} & \textbf{64.41} \\
\hline 
\hline
\end{tabular}
\end{center}
\caption{Evaluation on the tasks of amodal completion and ordering recovery. For comparison with \cite{zhan2020self}, we present the results of PCNet-m (reported) and  PCNet-m (reproduced), where the former results are reported in \cite{zhan2020self}, and the latter are obtained by retraining their public code from scratch. `-' indicates that results are not reported. 
}
\label{tab:variants}
\end{table*}

\textbf{Datasets}: We evaluate \Approach~on COCOA \cite{zhu2017semantic} and KINS \cite{qi2019amodal}, which are two benchmark amodal instance segmentation datasets with real images derived from the MSCOCO \cite{lin2014microsoft} and KITTI \cite{Geiger2012CVPR} datasets, respectively. COCOA consists of 2500, 1323, and 1250 images for training, validation, and test, respectively. There are 2140 object categories that can be divided into two superclasses: stuff (e.g., sky, grass, sea) and things (e.g.,  dog, cat, human). The content of images is mostly dense with multiple objects occluding one another in cluttered scenes. On the other hand, KINS is a large-scale traffic dataset, which consists of 7474 images for training and 7517 images for testing. There are 7 object categories in KINS including: cyclist, pedestrian, car, tram, truck, van, and miscellaneous vehicles. Scenes in KINS images are less cluttered than in COCOA, and if objects are occluded the occlusion is by mostly one other object.  We further randomly divide the KINS training set into training and validation sets with 6000 and 1474 images, respectively. 
Both datasets provide the ground-truth amodal masks, which we use only for evaluation.

\textbf{Evaluation Tasks and Metrics}: We evaluate \Approach~ on three tasks: ordering recovery, amodal completion, and amodal instance segmentation. 

For ordering recovery, we estimate the following relationships. Let $(\m_j, \m_j^A)$ and $(\m_k, \m_k^A)$ denote two pairs of input modal and output amodal masks of two \textit{adjacent} objects $j$ and $k$, respectively, where $|\m_j^A-\m_j|$ and  $|\m_k^A-\m_k|$ are the extended areas after amodal segmentation of $j$ and $k$. Then, we specify the ordering of $j$ and $k$ as%
\begin{equation}
    O(j,k)=\left\{\begin{array}{rl}
    0, & \text { if }\left|\m_j^A-\m_j\right|=\left|\m_k^A-\m_k\right|=0 \\
    1, & \text { if }\left|\m_j^A-\m_j\right|<\left|\m_k^A-\m_k\right| \\
    -1, & \text { otherwise },
\end{array}\right.
\end{equation}
where $O(j,k) = 1$ indicates that $j$ occludes $k$. If $j$ and $k$ are not adjacent, we set $O(j,k) = 0$. We evaluate our performance on the task of ordering recovery in terms of the average pairwise accuracy, O-Acc, between our predicted ordering relationships $O(j,k)$ and the ground truth relationships $O^*(j,k)$, for all pairs $(j,k)$ of adjacent objects.

For amodal completion, we compute the mean intersection-over-union, mIOU, between the predicted and ground-truth amodal masks, as well as invisible mIoU, inv-mIoU, for the predicted and ground-truth occluded regions. 

For amodal instance segmentation, we report the common metrics suggested by COCO, including average precision AP for thresholds 50\%, 75\%, 95\%, and average recall AR for top 1, 10, 100 predictions, among others.

We evaluate the following baseline and ablations: 
\begin{itemize}[itemsep=-2pt,topsep=-1pt, leftmargin=20pt]
    \item PCNet-m: our strong baseline from \cite{zhan2020self}.
    \item \mbox{Boundary$\to$PCNet-m}: in the input to PCNet-m we replace the occluder mask with the occlusion boundary; this tests only our contribution 1 (see Fig.~\ref{fig:main_diagram}), as PCNet-m does not estimate uncertainty.
    \item \mbox{Uncertainty$\to$PCNet-m}: PCNet-m is extended to predict uncertainty and trained with our uncertainty weighted loss, given by \eqref{eq:loss_function}, while the  input uses the occluder  mask as in \cite{zhan2020self}; this tests our 
    contribution 2 (see Fig.~\ref{fig:main_diagram}).
    \item \mbox{uBCE$\to$\Approach}: the uncertainty weighted loss, given by \eqref{eq:loss_function},  is replaced with the following uncertainty weighted binary cross-entropy (uBCE) loss for training our \Approach; this tests our proposed data term in \eqref{eq:loss_function}:
    \begin{equation}
    L_i^{\text{uBCE}} {=} \frac{1}{2} [-\frac{m_i^* \log m_i {+} (1{-} m_i^*) \log (1{-}m_i)}{u_i^2} {+} u_i^2].
\label{eq:uBCE_loss}
\end{equation} 
    \item \Approach: our full approach illustrated in Fig.~\ref{fig:main_diagram}.
    %
\end{itemize}

\subsection{Implementation Details}
\label{sec:implement}
Our implementation uses the github code of \cite{zhan2020self} as the base code and modify UNet \cite{ronneberger2015u} so it outputs two channels for amodal segmentation and uncertainty map, as described in Sec.~\ref{sec:loss}. We have also tested other networks for segmentation, such as DeepLabv3 \cite{chen2017rethinking} and DeepLabv3+ \cite{chen2018encoderdecoder}; however, our performance gain using these networks is statistically insignificant. We use the same training setting for fair comparison.  For learning, we use SGD with momentum \cite{cortes1995support}, and set the learning rate to $1e^{-4}$. The number of training iterations is 56000 and 32000 for COCOA and KINS, respectively. In each training iteration, we randomly choose Case 1 or Case 2 input data, as described in Sec.~\ref{sec:method}, to train \Approach~with a Bernoulli probability equal to $0.8$. The $\lambda$ in Eq.~\eqref{eq:loss_function} is set to 5 for the best performance. The threshold to binarize the amodal mask from our network's output is 0.5. The batch size for training UNet is 32 $256 \times 256$ images.


For amodal instance segmentation on test data, we use Mask-RCNN \cite{he2017mask} with ResNet50 \cite{he2015deep} as backbone and FPN \cite{lin2017feature} as the neck. The implementation of Mask-RCNN is provided in mmdetection \cite{chen2019mmdetection} toolbox. The batch size for training Mask-RCNN is 2 with the default setting provided by mmdetection.
All experiments are run on a PC with two 1080 Titan GPUs and 64 GB RAM.

\subsection{Ordering Recovery and Amodal Completion}
\label{sec:ordering}

Tab.~\ref{tab:variants} evaluates \Approach~on amodal completion and ordering recovery. For amodal completion, on COCOA-val, both \mbox{Boundary$\to$PCNet-m} and \mbox{Uncertainty$\to$PCNet-m} improve performance in mIoU over PCNet-m by $2.1\%$ and $1.8\%$ over PCNet-m (reproduced), respectively. A similar performance gain is observed on COCOA-test. Our contribution 1 (i.e., using the occlusion boundary mask in the input) has a larger effect on the performance than our contribution 2  (i.e., uncertainty), and each individual contribution leads to performance gains relative to the baseline.  The proposed integration of the two contributions in \Approach~gives the best amodal completion on both COCOA-val and COCOA-test. 

\begin{figure}[h!]
    \centering
    \includegraphics[scale=0.29]{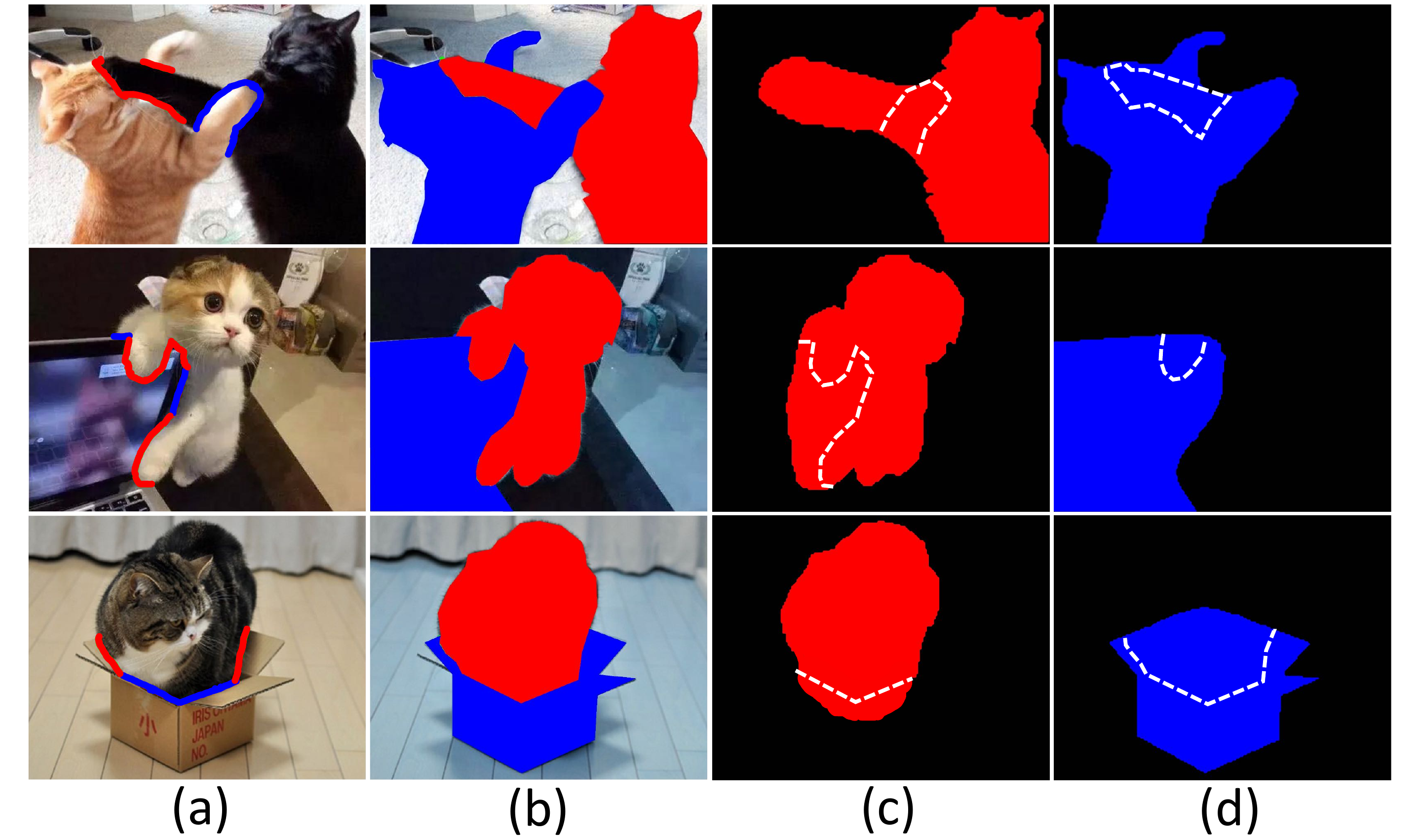}
    \caption{Mutual occlusion examples. \Approach~can successfully handle these cases while prior work \cite{zhan2020self} simply does not work since we cannot define the ordering graph. Each row is an example where red and blue colors represent objects A and B. For each row, from left to right: (a) input RGB image with color boundary segments indicating which object is in front of, (b) modal masks of the two objects, (c) and (d) \Approach~predicted amodal masks with white dash line indicating extended regions of objects A and B respectively.}
    \label{fig:mutual_occlusion}
\end{figure}

\begin{figure*}[h!]
    \centering
    \includegraphics[scale=0.45]{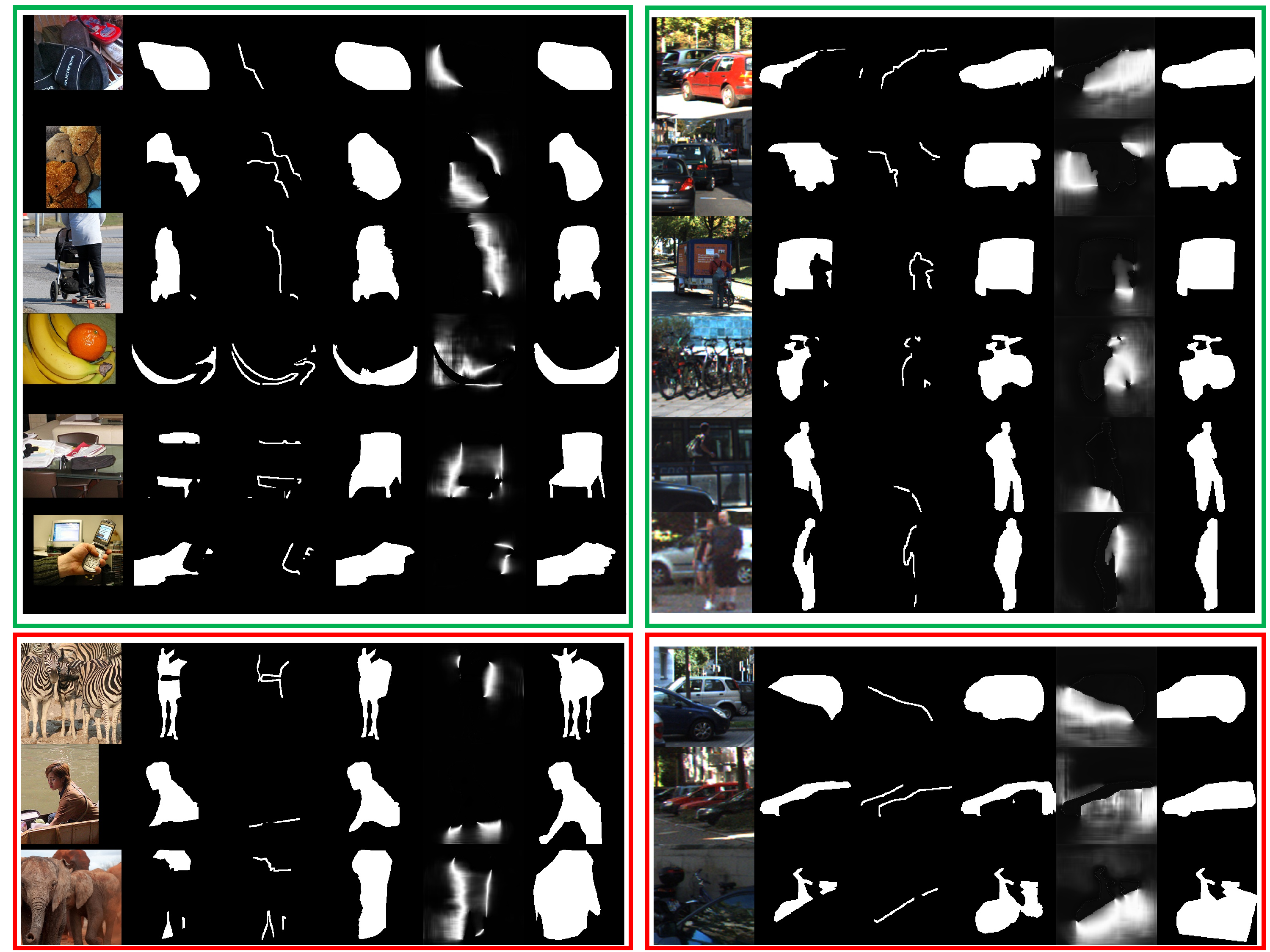}
    \caption{Qualitative results for amodal completion on COCOA-test (left) and KINS-test (right). For each column from left to right: (1) input RGB image, (2) input modal mask, (3) input occlusion boundary, (4) predicted amodal mask, (5) predicted uncertainty map, (6) GT amodal mask. Successful cases are in the green bounding boxes, and failure cases are in the red bounding boxes.}
    \label{fig:amodal_completion}
\end{figure*}

\begin{table*}[t]
\begin{center}
\small
\setlength{\tabcolsep}{5pt}
\begin{tabular}{ r  |l|ccc|ccc|ccc|ccc}
\hline
\hline
\textbf{Datasets} & \textbf{Trained on} & \textbf{AP} & \textbf{AP$_{50}$} & \textbf{AP$_{75}$} & \textbf{AP$_S$} & \textbf{AP$_M$} & \textbf{AP$_L$} & \textbf{AR$_{1}$} & \textbf{AR$_{10}$} & \textbf{AR$_{100}$} & \textbf{AR$_{S}$} & \textbf{AR$_{M}$} & \textbf{AR$_{L}$} \\
 \hline
\multirow{3}{*}{COCOA-val}  
& GT amodal       & 22.2 & 44.8 & 20.0 & 13.8 & 20.6 & 24.3 & 6.0 & 27.4 & 39.3 & 33.4 & 39.4 & 40.0 \\
& PCNet-m amodal  & 21.0 & 43.4 & 18.5 & 13.7 & 19.5 & 22.9 & 5.9 & 26.6 & 37.9 & 33.8 & 38.6 & 38.0 \\
& \Approach~amodal& 22.2 & 44.5 & 20.0 & 12.5 & 19.8 & 24.6 & 6.1 & 27.4 & 38.9 & 33.1 & 39.1 & 39.5 \\
\hline
\multirow{3}{*}{COCOA-test}  
& GT amodal       & 23.9 & 48.4 & 21.5 & 14.1 & 23.0 & 25.8 & 6.4 & 28.7 & 40.9 & 31.7 & 41.6 & 41.5 \\
& PCNet-m amodal  & 22.6 & 46.8 & 19.7 & 13.7 & 22.0 & 24.2 & 6.3 & 27.7 & 39.2 & 32.3 & 40.0 & 39.6 \\
& \Approach~amodal& 23.8 & 47.9 & 21.2 & 13.8 & 22.4 & 25.6 & 6.4 & 28.6 & 40.5 & 32.9 & 40.9 & 41.1 \\
\hline
\multirow{3}{*}{KINS-test}  
& GT amodal       & 30.8 & 53.9 & 31.6 & 15.1 & 40.4 & 56.7 & 18.9 & 38.3 & 40.4 & 24.1 & 51.6 & 65.6 \\
& PCNet-m amodal  & 29.1 & 51.8 & 29.6 & 14.1 & 38.1 & 55.7 & 18.3 & 37.1 & 38.9 & 23.0 & 49.4 & 65.2 \\
& \Approach~amodal& 29.3 & 52.1 & 29.7 & 14.2 & 38.2 & 56.0 & 18.4 & 37.0 & 38.8 & 23.1 & 49.3 & 64.9 \\
\hline
\hline
\end{tabular}
\end{center}
\caption{Amodal instance segmentation results of Mask-RCNN in full COCO metrics on COCOA-val, COCOA-test and KINS-test. Mask-RCNN is trained on either GT amodal masks, or PCNet-m generated amodal masks or \Approach~generated amodal masks.}
\label{tab:coco_segm}
\end{table*}

For KINS-test, on amodal completion, \Approach~ improves performance over PCNet-m (reproduced) in both mIoU and inv-mIoU. The performance gain in mIoU is relatively modest, which can be explained by certain properties of the dataset. KINS has fewer object categories than COCOA (7 vs. 2140),  and KINS scenes have fewer occlusions. Therefore, the ordering graph produced by PCNet-m is already highly accurate for amodal completion on KINS.

For the ordering recovery task on COCOA-val and COCOA-test, Tab.~\ref{tab:variants} shows a similar trend. \Approach~significantly outperforms PCNet-m (reproduced). 


Fig.~\ref{fig:mutual_occlusion} shows representative examples of two objects that mutually occlude each other. PCNet-m cannot handle such cases since their ordering graph is not expressive enough. On the contrary, \Approach~successfully infers the amodal masks of the two mutually occluding objects.

Fig.~\ref{fig:amodal_completion} illustrates results of \Approach~on COCOA and KINS on the task of amodal completion. As can be seen, \Approach~gives highly accurate predictions. The figure also shows our estimated uncertainty maps which usually take high values on object boundaries. In some cases,  \Approach~fails to fully complete the amodal masks, due to the high similarity of foreground and background.

\begin{figure*}[h!]
    \centering
    \includegraphics[scale=0.45]{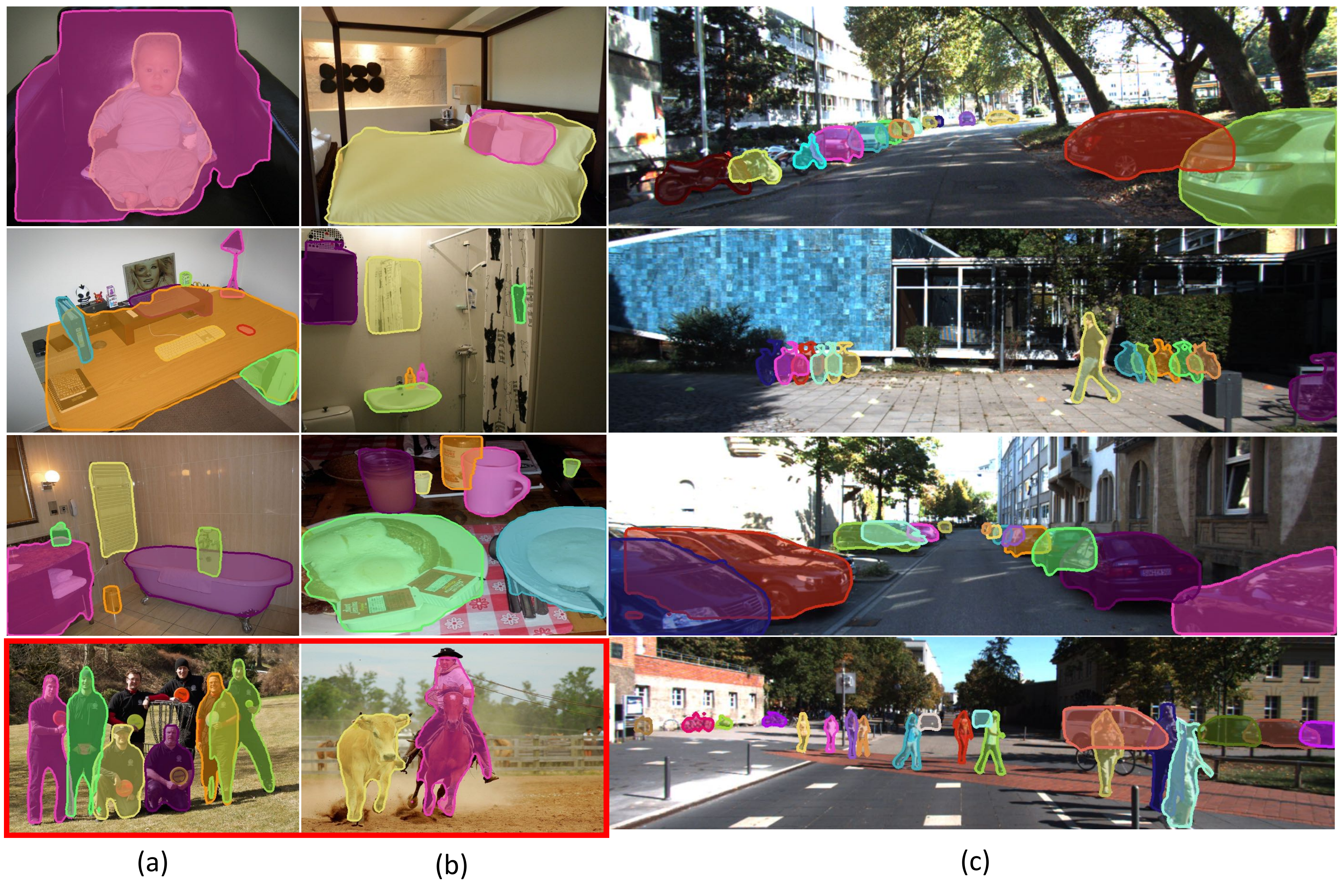}
    \caption{Qualitative results for amodal instance segmentation on COCOA-test are shown in (a) and (b), and on  KINS-test in (c). All of them are successful cases except the failure cases are marked red. The first failure case is about incomplete person detections and the second failure case is about merging masks of a person and horse.}
    \label{fig:amodal_instance_segmentation}
\end{figure*}

\subsection{Amodal Instance Segmentation}

For this task, we take the trained \Approach~to predict amodal masks on COCOA-train and KINS-train, and use these pseudo amodal masks to train Mask-RCNN (pre-trained on COCO) in 12 epochs (1x configuration) to predict amodal instance segmentation for COCOA-val, COCOA-test, and KINS-test. We use the evaluation code from the COCO dataset Github. We repeat the same process for trained PCNet-m. Because the number of classes in COCOA is too large (2140 classes) and our focus is on the quality of amodal segmentation, we group them into one foreground class to train and evaluate. For KINS, we keep the number of classes as specified in this dataset.

For reporting an upper-bound performance, we train Mask-RCNN on the ground-truth amodal masks of COCO-train and KINS-train to predict amodal instance segmentation of COCOA-val, COCOA-test, and KINS-test.

Tab.~\ref{tab:coco_segm} evaluates amodal instance segmentation using Mask-RCNN trained on: ground-truth amodal segmentations  (GT amodal), and pseudo-ground truth produced by PCNet-m (PCNet-m amodal) and \Approach~(\Approach~amodal). From the table, on COCOA-val, a difference in AP between GT amodal and \Approach~amodal is zero. On COCOA-test, when using the pseudo-ground truth from \Approach, we increase AP relative to that of PCNet-m amodal. Tab.~\ref{tab:coco_segm} suggests that on COCOA \Approach~pseudo amodal masks have similar quality as the actual ground truth. 
On KINS-test, ~\Approach~amodal gives slightly better results than PCNet-m amodal (with a 0.2 margin) while we are behind from GT amodal by 1.5 in AP. This can be explained in terms of simpler scenes in KINS relative to those in COCOA.

Fig.~\ref{fig:amodal_instance_segmentation} shows representative results of Mask-RCNN trained with \Approach's pseudo amodal masks. For the COCOA-test dataset, we usually obtain good results, with some exceptions due to the problems of incomplete instance modal segmentation. Also, on the KINS-test dataset, we obtain very good amodal instance segmentations.

\section{Conclusion}
\label{sec:conclusions}

We have specified a new amodal segmenter with boundary uncertainty estimation  (\Approach) for weakly supervised amodal instance segmentation. To address the lack of ground-truth amodal masks, we have trained \Approach~on manipulated images to produce pseudo-ground truth amodal masks, and then learned a common instance segmenter, Mask-RCNN, on our pseudo-ground truth. We have made two contributions. First, we have replaced the occluder mask used in prior work \cite{zhan2020self} for input with the occlusion boundary, and consequently removed the need for one step in \cite{zhan2020self} -- that of estimating the object ordering graph. Second, we have enabled \Approach~to estimate uncertainty of the predicted amodal segmentation and proposed a new loss function that uses the estimated uncertainty to regularize learning of \Approach. Our evaluation on the tasks of amodal completion, ordering recovery, and amodal instance segmentation, on the COCOA dataset, demonstrates that \Approach~outperforms the state of the art.
Specifically, in comparison with a strong baseline PCNet-m \cite{zhan2020self}, our performance improves by $3.5\%$ and $4.5\%$ in mean intersection-over-union (mIoU) for amodal completion and average pairwise accuracy (O-Acc) for ordering recovery, respectively. In amodal instance segmentation, Mask-RCNN trained on our pseudo amodal masks has nearly the same performance as Mask-RCNN trained on the ground-truth amodal masks with a performance gap of $0.1$ in average precision (AP) on the COCOA-test dataset. 

{\bf Acknowledgement.} This work was supported in part by  DARPA MCS Award N66001-19-2-4035.

{\small
\bibliographystyle{ieee_fullname}
\bibliography{egbib}
}

\end{document}